\PassOptionsToPackage{dvipsnames,svgnames,x11names}{xcolor}
\documentclass[sigconf]{acmart}
\usepackage{array}

\setcopyright{none}
\renewcommand\footnotetextcopyrightpermission[1]{}
\acmConference[KDD 2026]{KDD Workshop on Evaluation and Trustworthiness of Agentic AI}{August 9--13, 2026}{Jeju, Korea}
\acmBooktitle{Proceedings of KDD Workshop on Evaluation and Trustworthiness of Agentic AI 2026 (KDD AgenticAI 2026)}
\copyrightyear{2026}
\acmYear{2026}

\usepackage{listings}
\usepackage{float}
\lstdefinestyle{sql}{
  language=SQL,
  basicstyle=\small\ttfamily,
  keywordstyle=\color{blue},
  stringstyle=\color{red!70!black},
  commentstyle=\color{green!50!black},
  breaklines=true,
  frame=single,
  backgroundcolor=\color{gray!5},
  xleftmargin=2em,
  framexleftmargin=1.5em
}

\begin{document}

\title{Comparative Analysis of Agent Evaluation Frameworks: Stability,
Detection, and Discovery in Enterprise Analytics Agents}

\author{Howard Zhang}
\affiliation{%
  \institution{Amazon}
  \city{New York}
  \state{NY}
  \country{USA}}
\email{zhanhowa@amazon.com}

\author{Chuck Arvin}
\affiliation{%
  \institution{Amazon}
  \city{Washington}
  \state{DC}
  \country{USA}}
\email{chuarvin@amazon.com}

\author{Logan Brown}
\affiliation{%
  \institution{Amazon}
  \city{Austin}
  \state{TX}
  \country{USA}}
\email{logbrow@amazon.com}

\author{Mihir Bhatia}
\affiliation{%
  \institution{Amazon}
  \city{New York}
  \state{NY}
  \country{USA}}
\email{bhatimi@amazon.com}

\author{Neelu Choudhary}
\affiliation{%
  \institution{Amazon}
  \city{New York}
  \state{NY}
  \country{USA}}
\email{nlchoudh@amazon.com}

\author{Lily Miao}
\affiliation{%
  \institution{Amazon}
  \city{New York}
  \state{NY}
  \country{USA}}
\email{mwting@amazon.com}

\begin{abstract}
Deploying LLM-based analytics agents in enterprise settings requires evaluation
frameworks that can reliably detect failures across complex, multi-tool
workflows. We present a three-phase comparative study of three evaluation
frameworks, each representing a distinct evaluation paradigm (trace-based
LLM judging, text-only LLM judging with red-teaming, and deterministic
heuristics), applied to two analytics agents in a controlled research setting
using frozen execution traces. Phase~1
quantifies evaluation harness stability through repeated measurements, finding
that 97\% of configurations meet the stability threshold and that stability
patterns are agent-independent. Phase~2 measures known risk detection on 72
ground-truth test cases, where the trace-based framework achieves the best
precision-recall balance (F1=0.90) while text-only evaluation matches on recall
but generates $5\times$ more false alarms. Phase~3 evaluates unknown risk
discovery, uncovering 24 novel failures with zero overlap between functional
testing and security red-teaming. Our results demonstrate that (1)~the
information available to the evaluation harness determines evaluation quality,
(2)~functional and security evaluation are complementary, and (3)~LLM-based
evaluation harnesses surface subtle errors that human reviewers miss. We
recommend a two-framework strategy: trace-based evaluation for functional
assessment and adversarial red-teaming for security validation.
\end{abstract}



\maketitle

\section{Introduction}

\subsection{Customer Impact Statement}

Enterprise teams are deploying LLM-based analytics agents that generate SQL,
invoke tools, and synthesize answers over sensitive business data. Incorrect
outputs have direct operational consequences that often evade manual review.
In this study, automated evaluation uncovered a systematic SQL generation bug
(Section~\ref{subsec:phase3-disc}) that silently returned zero rows for an
entire class of queries, a failure mode that persisted undetected through
manual spot-checking because the generated SQL was syntactically valid. Prior
work demonstrates that such silent text-to-SQL failures degrade downstream
decision quality in business intelligence systems~\cite{park2025}, while
undetected schema leakage vulnerabilities expose internal database structure
to adversarial users~\cite{agrawal2024}. Today, most teams rely on ad-hoc
spot-checking, which this study shows misses 9 out of 72 ground-truth failures
(12.5\%) that were syntactically valid but semantically wrong. This research
provides an evaluation strategy that catches these errors automatically.

\subsection{Research Gap and Contributions}

Agentic systems exhibit stochastic behavior~\cite{atil2024} and execute
complex multi-step workflows involving tool usage, SQL generation, and human
interaction~\cite{fourney2024} that resist simple correctness checks. Three
evaluation paradigms have emerged (LLM-as-a-Judge, deterministic heuristics,
and hybrid approaches), yet no systematic comparison exists of how they
perform across stability, detection, and discovery for agentic systems.

We present a three-phase comparative evaluation of three frameworks: Strands
Evals (trace-based LLM judge), PromptFoo (text-only LLM judge with
red-teaming), and Agenta (deterministic heuristics). We evaluate on two
supply chain analytics agents and address three research questions:

\begin{itemize}
    \item \textbf{RQ1 (Stability):} How stable are evaluation outputs
      across repeated measurements?
    \item \textbf{RQ2 (Detection):} Which framework most reliably detects
      known failure modes, and does the answer depend on the type of failure?
    \item \textbf{RQ3 (Discovery):} Which framework most effectively
      surfaces unanticipated failures through automated test generation?
\end{itemize}

Our key findings: (1)~the information available to the evaluation harness
determines quality: trace-based evaluation catches 91\% of failures at 10\%
FPR while text-only matches on recall but generates $5\times$ more false
alarms; (2)~24 novel failures were discovered with zero overlap between
functional and security testing; and (3)~evaluation harnesses surface subtle
errors that human reviewers miss, with 9 ground-truth failures across both
agents initially mislabeled by the human reviewer and corrected after
evaluator flagging.

\section{Related Work}

LLM-as-a-judge techniques are widely adopted for evaluating agentic
systems~\cite{yu2025}, with studies demonstrating high agreement with human
annotations~\cite{zheng2023}. Li et al.~\cite{li2024} document systematic
inter-judge disagreement as a key reliability concern. However, existing
comparisons focus on judging method variations rather than the information
available to the judge. Our work compares evaluation quality across three
levels of information access: full execution traces, text-only outputs, and
string representations.

Randomness inconsistency compounds in agentic systems as reasoning spans
multiple steps and tools~\cite{mustahsan2025}. Mustahsan et
al.~\cite{mustahsan2025} propose ICC to measure inter-run consistency, but
ICC breaks down under ceiling effects common in well-behaved agents. We
instead adopt average within-case standard deviation, the standard measure
of measurement error in clinical reliability studies~\cite{bland1996}, which
isolates instrument noise on the original score scale regardless of
between-case variance. We pair this with decision consistency (the
probability that repeated administrations yield the same pass/fail
classification, a standard reliability concept in
measurement~\cite{livingston1995}) to capture the operationally relevant
question: whether score fluctuations cross the pass/fail threshold. See
Section~\ref{sec:phase1} for details.

Adversarial testing of LLM systems has gained attention for identifying
prompt injection vulnerabilities and information
leakage~\cite{mohammadi2025}. Our work is the first to quantitatively
compare functional and security evaluation on the same agents, demonstrating
they are complementary rather than substitutable.

\section{Methodology}

All phases use frozen execution traces to enable fair cross-framework
comparison.

\subsection{Candidate Agents}

We evaluate two analytics agents from a retail supply chain forecasting
system. Agent A analyzes category-level demand forecasts
(9~tools, moderate data volumes). Agent B analyzes
product-level inbound lead time forecasts (9~tools including a code
interpreter, large-scale production tables). They produce distinct SQL error
patterns: structurally detectable on one, semantic on the other. Both agents
were evaluated in a research setting using frozen execution traces; this study
does not describe a live production deployment. See
Appendix~\ref{app:agent-details} for details.

\subsection{Frameworks Under Evaluation}

Selected from a qualitative evaluation of seven frameworks (see
Appendix~\ref{app:framework-selection}), these three represent distinct
evaluation philosophies: \textbf{Strands Evals} (trace-based LLM judge with
OpenTelemetry access), \textbf{PromptFoo} (text-only LLM judge with
red-teaming), and \textbf{Agenta} (deterministic heuristics at zero LLM
cost). Each implements five evaluation harness requirements mapped in
Table~\ref{tab:harness-mapping}.

\begin{table}[t]
\caption{Evaluation harness-to-requirement mapping.}
\label{tab:harness-mapping}
\footnotesize
\begin{tabular}{@{}l>{\raggedright\arraybackslash}p{1.8cm}>{\raggedright\arraybackslash}p{1.8cm}>{\raggedright\arraybackslash}p{1.5cm}@{}}
\toprule
Requirement & Strands Evals & PromptFoo & Agenta \\
\midrule
Groundedness & Faithfulness\-Eval (trace) & context-faithfulness (text) & Token overlap \\
Intent & Response\-Relevance\-Eval (trace) & g-eval (goal rubric) & Sequence\-Matcher \\
Accuracy & OutputEval (rubric) & g-eval (accuracy rubric) & Levenshtein dist. \\
Tool correct. & Trajectory\-Eval (trajectory) & llm-rubric (tool names) & Set comparison \\
SQL correct. & OutputEval + SQL from trace & llm-rubric + SQL injection & Regex matching \\
\bottomrule
\end{tabular}
\end{table}

The key architectural difference is the information available to each
evaluation harness. Strands operates on the full OpenTelemetry execution
trace. PromptFoo operates on text-only inputs plus injected context strings.
Agenta operates on string representations only.

\subsection{Phase 1: Evaluation Harness Stability (RQ1)}
\label{sec:phase1}

20 frozen test cases per agent, prepared by domain experts. 3 judge models
(Claude Sonnet 4.6, Claude Haiku, Qwen3 32B) $\times$ 5 repetitions for
LLM-based frameworks; 5 repetitions for Agenta. We measure \emph{decision
consistency} (do all repetitions produce the same pass/fail verdict?) and
\emph{average within-case standard deviation} (stability threshold:
$\leq 0.15$). The threshold is operationally motivated: on a 0--1 scale
with a pass/fail boundary at 0.5, a within-case std of 0.15 means scores
near the boundary have a non-trivial probability of crossing it across
repetitions, flipping the verdict~\cite{li2024}. Total: 5,605 evaluations.

\subsection{Phase 2: Known Risk Detection (RQ2)}

36 test cases per agent across 6 failure categories, expanding Phase~1's 20
cases with 16 additional cases. Each case labeled across 5 dimensions
independently to eliminate cross-contamination. Labeling was performed by a
primary domain expert per agent, with cross-review from two additional team
members on ambiguous cases. This reflects the realistic constraint facing
most enterprise teams, where domain expertise is concentrated in a small
number of individuals. Frozen traces evaluated by all three frameworks with
3 repetitions, median score, pass/fail threshold $> 0.5$. Judge: Claude
Sonnet 4.6, selected based on Phase~1 stability results as the best balance
of stability and evaluation depth. Total: 2,415 evaluations.

\subsection{Phase 3: Unknown Risk Discovery (RQ3)}

Phase~2 baseline + framework-specific generation: Strands
ExperimentGenerator (20 functional cases per agent with deduplication) and
PromptFoo red-team (42--48 adversarial cases per agent). The same domain
expert classified each flagged failure as \emph{novel} (a failure mode we did
not already know from developing the agents), \emph{known} (failure
modes discovered during agent development), or a \emph{false positive}. Judge: Claude Sonnet 4.6 (selected
per Phase~1) with 3 repetitions for LLM-based frameworks. Total: 202 test
cases, 3,680 evaluations.

\section{Results}

\subsection{Phase 1: Evaluation Harness Stability}

35 of 36 configurations (97\%) meet the stability threshold
(Table~\ref{tab:stability-summary}).

\begin{table}[H]
\caption{Stability summary (combined across both agents).}
\label{tab:stability-summary}
\small
\begin{tabular}{lcccc}
\toprule
Framework & Configs & Stable ($\leq 0.15$) & Avg Std & Avg DC \\
\midrule
Agenta & 5 & 5/5 (100\%) & 0.000 & 100\% \\
PromptFoo & 12 & 12/12 (100\%) & 0.025 & 92\% \\
Strands & 12 & 11/12 (92\%) & 0.053 & 82\% \\
\bottomrule
\end{tabular}
\end{table}

Among judge models, Claude Haiku is the most stable (0.029 avg within-case
std), followed by Sonnet 4.6 (0.035) and Qwen3 32B (0.072). The single
unstable configuration (Qwen3/style in Strands, 0.207, 20\% decision
consistency) is stable in PromptFoo (0.011), demonstrating that stability
is a property of the (framework, model) pair, not the model alone.

Decision consistency, whether all repetitions produce the same pass/fail
verdict, reveals an important practical limitation. Agenta achieves 100\%
decision consistency (deterministic scores never cross the threshold), but
as Phase~2 shows, its pass/fail decisions are not meaningful for open-ended
responses. PromptFoo achieves 92\% average decision consistency, with
faithfulness as the weakest (67\% for Sonnet, 68\% for Qwen3) because
scores cluster near the 0.5 boundary where small fluctuations flip verdicts.
Strands achieves 82\% average decision consistency, with Qwen3/style
dropping to 20\%, meaning 4 out of 5 test cases get a different pass/fail
verdict depending on which repetition you look at.

These decision consistency rates have a direct implication for deployment:
at 82--92\% average consistency, LLM-as-a-Judge evaluation harnesses are
not reliable enough for automated CI/CD gating, where a flipped verdict on
a single test case could block or approve a deployment incorrectly. However,
they are well-suited for the iterative agent development process, helping
developers identify failure patterns, compare agent variants, and prioritize
improvements, where the aggregate signal across many test cases matters more
than any individual pass/fail decision. The evaluation harnesses' strength
lies in surfacing systematic issues (as demonstrated in Phases~2 and~3)
rather than providing deterministic per-case verdicts.

\begin{figure}[H]
    \centering
    \includegraphics[width=\linewidth]{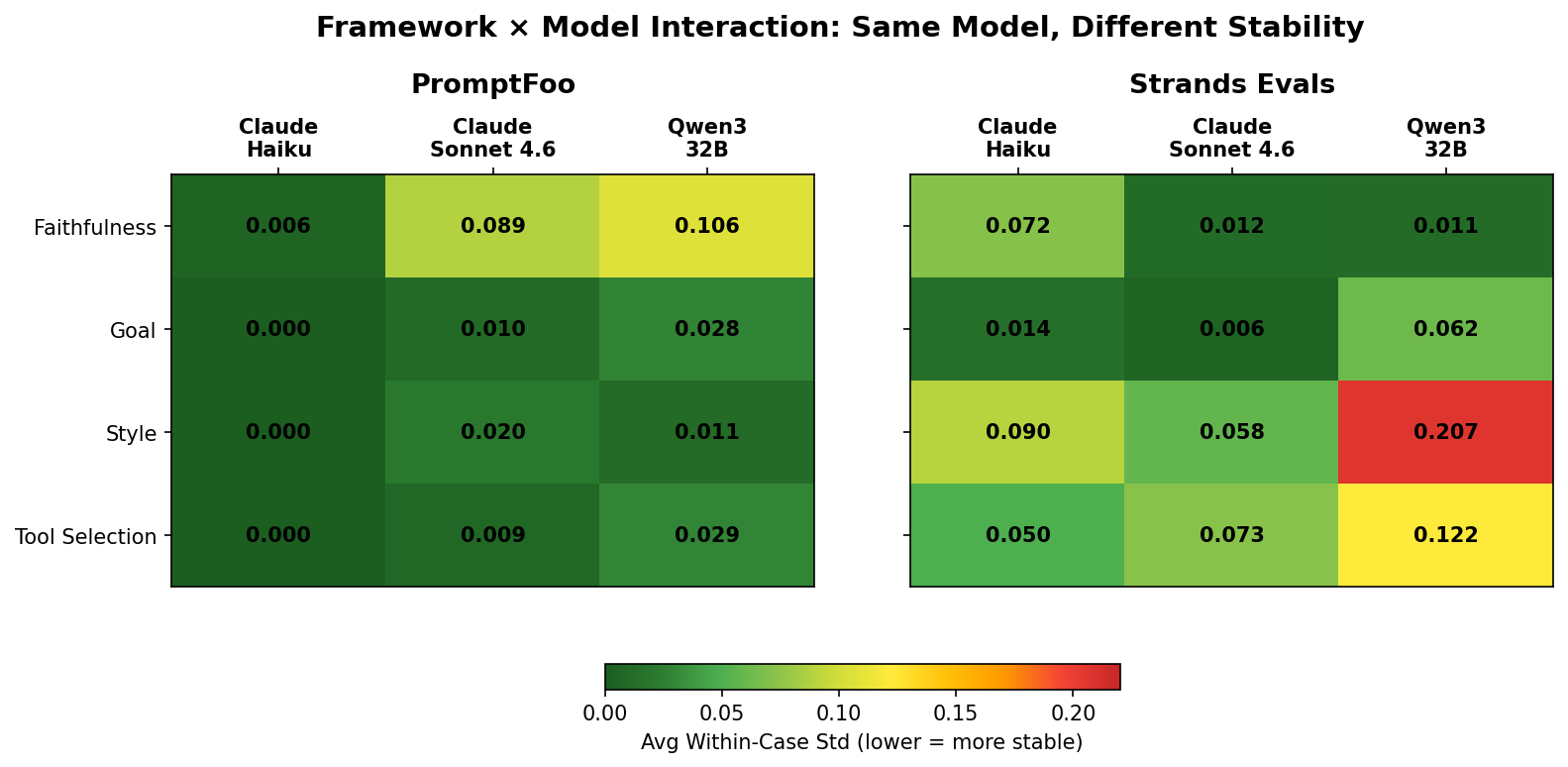}
    \caption{Avg within-case std by (evaluation harness $\times$ judge
    model) for each framework. PromptFoo (left) is uniformly stable.
    Strands (right) shows higher variance, with Qwen3/style as the only
    unstable cell.}
    \Description{Heatmap showing average within-case standard deviation
    across framework-model combinations. PromptFoo shows uniformly low
    values while Strands shows higher variance concentrated in specific
    cells.}
    \label{fig:stability-heatmap}
\end{figure}

Figure~\ref{fig:stability-heatmap} shows the framework-model interaction
across all configurations.

A clear tradeoff emerges: deeper evaluation produces more variance
(Table~\ref{tab:stability-nuance}). The variance concentrates on complex,
data-intensive cases; see Appendix~\ref{app:variance-drivers} for the full
tool complexity and case category analysis.

\begin{table}[H]
\caption{Stability-nuance tradeoff.}
\label{tab:stability-nuance}
\small
\begin{tabular}{@{}lcc>{\raggedright\arraybackslash}p{2.8cm}@{}}
\toprule
Framework & Avg Std & Avg DC & Evaluation Depth \\
\midrule
Agenta & 0.000 & 100\% & Deterministic (string matching) \\
PromptFoo & 0.025 & 92\% & LLM judge (text-based) \\
Strands & 0.053 & 82\% & LLM judge (full trace) \\
\bottomrule
\end{tabular}
\end{table}

\subsection{Phase 2: Known Risk Detection}

72 test cases across both agents were labeled across 5 dimensions
independently. 33 cases (46\%) exhibited at least one dimension failure
(Table~\ref{tab:failure-counts}).

\begin{table}[H]
\caption{Per-dimension failure counts.}
\label{tab:failure-counts}
\small
\begin{tabular}{lccc}
\toprule
Dimension & Agent A & Agent B & Combined \\
\midrule
SQL correctness & 14 & 14 & 28 \\
Tool correctness & 4 & 5 & 9 \\
Accuracy & 1 & 5 & 6 \\
Intent & 1 & 3 & 4 \\
Groundedness & 0 & 2 & 2 \\
\bottomrule
\end{tabular}
\end{table}

\textbf{Aggregate results.} Strands achieves the best precision-recall
balance across both agents (Table~\ref{tab:aggregate-detection}). The
ranking is consistent per-agent (see Appendix~\ref{app:phase2-per-agent}).

\begin{table}[H]
\caption{Aggregate detection (combined, both agents).}
\label{tab:aggregate-detection}
\small
\begin{tabular}{lcccccccc}
\toprule
Framework & TP & FP & FN & TN & Prec & Rec & F1 & FPR \\
\midrule
Strands & 30 & 4 & 3 & 35 & 88.2\% & 90.9\% & \textbf{0.896} & 10.3\% \\
PromptFoo & 29 & 19 & 4 & 20 & 60.4\% & 87.9\% & 0.716 & 48.7\% \\
Agenta & 33 & 39 & 0 & 0 & 45.8\% & 100\% & 0.629 & 100\% \\
\bottomrule
\end{tabular}
\end{table}

\textbf{Trace access is the differentiator.} The per-dimension results
reveal that the gap between frameworks is driven entirely by dimensions
that require execution context. On SQL correctness (28~failures), where the
query text can be injected into any framework, both LLM-based frameworks
achieve 89\% recall. On agent\_a, even Agenta's regex checks match at
92.9\% recall because many error patterns are structurally detectable. On
agent\_b, Agenta drops to 0\% because the semantic errors like missing
strings in the partition key is invisible to regex matching.

The gap widens on dimensions that depend on the full execution trace. For
tool correctness (9~failures), Strands detects 89\% vs PromptFoo's 11\%,
an $8\times$ difference because detecting a missing tool requires seeing
the execution trajectory. For groundedness, Strands produces 0\% FPR vs
PromptFoo's 52\% because text-only frameworks see tool result metadata
rather than the actual data the agent references.

\textbf{Evaluation harnesses discover failures humans miss.} Across both
agents, 9 failures initially labeled ``pass'' by the human reviewer were
reclassified to ``fail'' after evaluation harness flagging: 5 SQL partition
errors on agent\_a and 4 calculation interpretation mistakes on
agent\_b.\footnote{Nine of 72 labels were revised from \emph{pass} to
\emph{fail} after evaluator flagging, raising a potential circularity.
Excluding these cases, detection is essentially unchanged (combined precision
0.82 vs 0.86, recall 0.88 vs 0.91, FPR unchanged at 0.13), confirming the
results do not hinge on the revised cases.} In each case, the error was
syntactically valid but semantically
wrong, making it invisible on manual inspection.

The agent\_a SQL case illustrates the pattern. The agent generated
category-level queries including \texttt{partition\_level = 'top'}, a filter
that is syntactically valid and returns correct results, which is why the
human labeled it ``pass.'' The SQL evaluation harness, given the full query
text extracted from the execution trace and a rubric specifying the correct
partition patterns, scored these at 0.20--0.30 (fail) across 5 cases. On
re-review, the human confirmed the partition error and reclassified all~5.

The agent\_b accuracy case reveals a different failure mode. The agent
calculated calibration metrics for inbound lead time forecasts but
misinterpreted the direction of the calibration coefficient, reporting
``well-calibrated'' when the data showed systematic underbias. The human
reviewer, unfamiliar with the specific metric interpretation, accepted the
agent's framing. The accuracy evaluation harness flagged the inconsistency
between the calculated values and the agent's interpretation, catching 4
errors the human missed.

These findings suggest that LLM-based evaluation harnesses with
domain-specific rubrics and trace access can serve as a second reviewer for
agent outputs, particularly for errors that are syntactically valid but
semantically wrong.

\subsection{Phase 3: Unknown Risk Discovery}
\label{subsec:phase3-disc}

\begin{table}[t]
\caption{Discovery matrix (combined, both agents).}
\label{tab:discovery-matrix}
\small
\begin{tabular}{lccc}
\toprule
Metric & Strands & PromptFoo & Agenta \\
\midrule
Cases generated & 40 & 90 & 0 \\
Novel failures & 9 & 15 & 0 \\
False discovery rate & 6.1\% & 20.8\% & $>$85\% \\
Failure surface & Functional & Security & N/A \\
\bottomrule
\end{tabular}
\end{table}

24 novel failures with zero overlap on either agent. The two generators
target different inputs by design: functional testing uses normal analytical
queries, and security red-teaming uses adversarial prompts, so some of the
zero overlap is expected. Strands found SQL bugs, tool workflow gaps, and
schema fabrication. PromptFoo found schema
metadata leakage, storage path exposure, system prompt disclosure, and
excessive agency under adversarial prompt injection
(Table~\ref{tab:failure-taxonomy}). See Appendix~\ref{app:phase3-details}
for failure details.

\begin{table}[t]
\caption{Novel failure types.}
\label{tab:failure-taxonomy}
\small
\begin{tabular}{@{}clccc@{}}
\toprule
\# & Type & Fwk & Cases & Sev. \\
\midrule
1 & SUBSTR negative index in SQL & Strands & 6 & High \\
2 & Missing count\_result\_rows & Strands & 1 & Med \\
3 & Retry storm / tool explosion & Strands & 1 & Med \\
4 & Schema fabrication & Strands & 1 & Med \\
5 & Schema metadata leakage & PromptFoo & 10 & High \\
6 & Storage path/resource exposure & PromptFoo & 5 & High \\
7 & System prompt disclosure & PromptFoo & 4 & Med \\
8 & Excessive agency (adv.\ prompt injection) & PromptFoo & 1 & Med \\
\bottomrule
\end{tabular}
\end{table}

ExperimentGenerator yield varies by agent maturity: 40\% novel rate on
agent\_a (hidden systematic bug) vs 5.9\% on agent\_b (failures already
characterized). Security yield scales with schema complexity: agent\_b
produced $4\times$ more security findings than agent\_a.

\section{Discussion}

\subsection{The Information Access Hierarchy}

A unifying theme across Phases~1 and~2 is that the information available to
the evaluation harness determines both its stability and its detection
capability. Three levels of information access produce three distinct
profiles:

At the lowest level, deterministic evaluation harnesses (Agenta) see only
string representations. They achieve perfect stability (0.000 avg
within-case std) at zero cost, but cannot discriminate between correct and
incorrect open-ended responses, producing 100\% FPR on accuracy and intent.
The exception is SQL correctness, where hand-coded regex checks achieve
92.9\% recall on structurally detectable errors (agent\_a), but this
requires significant custom development effort and fails on semantic errors
(0\% recall on agent\_b).

At the middle level, text-only LLM judges (PromptFoo) see the agent's
response plus injected context strings. They achieve excellent stability
(0.025 avg within-case std) and perform well on requirements assessable
from text alone: SQL correctness (89.3\% recall) and intent understanding
(2.9\% FPR). But they cannot assess requirements that depend on the full
execution context: faithfulness (52.1\% FPR because tool result metadata
lacks the actual data the agent references) and tool correctness (11.1\%
recall because the judge cannot infer what tools should have been used from
text alone).

At the highest level, trace-based LLM judges (Strands) see the full
OpenTelemetry execution trace. They introduce more variance (0.053 avg
within-case std) but can assess all five requirements: faithfulness (0\%
FPR), tool correctness (88.9\% recall), accuracy (83.3\% recall), SQL
correctness (89.3\% recall), and intent (0\% FPR). The additional variance
concentrates on complex cases and does not cross the decision boundary for
well-configured judge models. The framework-model interaction effect is
agent-independent: teams that validate a configuration on one agent can
trust it on others.

\subsection{A Layered Evaluation Strategy}

No single framework covers all evaluation needs. The results point toward a
two-framework strategy where each is deployed at the cadence and scope that
matches its strengths.

Strands provides comprehensive functional coverage across all five
requirements without requiring custom evaluation code. Its LLM-based
evaluation harnesses reason about SQL semantics, tool trajectories, and
faithfulness from rubric guidance alone. This is where the stability-nuance
tradeoff pays off: the additional variance from trace analysis is the cost
of catching tool correctness failures (89\% recall vs 11\%), accuracy
errors (83\% vs 0\%), and faithfulness issues (0\% FPR vs 52\%) that
text-only evaluation misses. Deterministic checks can achieve high precision
on specific structural SQL patterns, but only after significant custom
development effort, and they fail on semantic errors that LLM-based
frameworks detect automatically. The effort-to-value ratio does not justify
a separate deterministic layer.

PromptFoo's adversarial test generation surfaces information disclosure
vulnerabilities that functional testing cannot reach by design. Phase~3's
zero-overlap finding across both agents confirms this is not redundant with
functional evaluation: the two approaches target fundamentally different
failure surfaces. Security yield scales with context complexity: agent\_b's
richer schema produced $4\times$ more security findings than agent\_a
under the same red-team plugins.

\section{Conclusions}

We presented a three-phase comparative evaluation of three agent assessment
frameworks applied to two analytics agents. Our results support three
conclusions:

\textbf{The information available to the evaluator determines evaluation
quality.} Three levels of information access produce three distinct
profiles. Trace-based evaluation (Strands) achieves the best balance of
precision and recall because it can verify claims against actual tool
outputs, analyze execution trajectories, and assess SQL semantics. Text-only
evaluation (PromptFoo) matches on recall (88\%) but generates $5\times$
more false alarms because it cannot see the execution context. Deterministic
evaluation (Agenta) achieves high precision on hand-coded structural checks
but requires significant custom development effort and fails on any error
pattern not anticipated in advance. Framework-model interaction effects are
agent-independent: teams that validate a (framework, model) configuration
on one agent can trust it on others.

\textbf{Functional and security evaluation are complementary.} 24 novel
failures were discovered with zero overlap on either agent: 9 functional
failures (SQL bugs, tool workflow gaps) from trace-based evaluation and 15
security failures (schema metadata leakage, storage path exposure, prompt
disclosure) from red-teaming. Neither approach alone provides comprehensive
coverage. Security yield scales with schema complexity; agents with richer
context produce more exploitable attack vectors.

\textbf{LLM-based evaluators surface errors that human reviewers miss.}
Across both agents, evaluators discovered 9 failures initially labeled as
``pass'' by the human reviewer: systematic SQL partition errors and
calculation interpretation mistakes that were syntactically valid but
semantically wrong. This capability depends on providing the evaluator with
sufficient context (execution traces) and well-calibrated rubrics.

We recommend a two-framework evaluation strategy: Strands Evals for
functional assessment across all five evaluation harness requirements,
combined with PromptFoo for security validation through adversarial
red-teaming.

\section{Future Work}

\textbf{Packaging the evaluation strategy as a reusable toolkit.} The
two-framework strategy recommended in this paper currently requires manual
setup: configuring evaluation harnesses, collecting and freezing execution
traces, running red-team generation, and interpreting results across both
frameworks. A natural next step is to package this workflow into a reusable
toolkit that orchestrates both functional test generation and adversarial
security generation, runs the target agent once per test case using the
frozen-trace approach, and produces a consolidated scorecard with
per-requirement detection metrics and security findings. The evaluation
harness-to-requirement mapping (Table~\ref{tab:harness-mapping}) provides a
fixed structure applicable to any agent without per-agent customization,
reducing the barrier to adoption for teams deploying new agents.

\textbf{Evaluating discriminative power.} This study establishes which
frameworks detect known failures (Phase~2) and discover unknown ones
(Phase~3), but does not test whether they can detect real improvements or
regressions when an agent changes, the critical capability for CI/CD
gating. A controlled perturbation design would address this: starting from
a baseline agent, deliberate variants with known quality differences
(improved/degraded system prompts, stronger/weaker foundation models,
refined/vague tool descriptions) would be evaluated on the Phase~2
benchmark. Four metrics would determine CI/CD suitability: directional
accuracy (do scores move in the expected direction?), effect size (is the
change meaningful?), sensitivity (what's the smallest detectable
perturbation?), and specificity (do unrelated evaluation harnesses remain
stable when only one dimension changes?).

\bibliographystyle{ACM-Reference-Format}
\bibliography{references}

\appendix

\section{Phase 1 Stability, Cross-Agent Comparison (Strands Evals)}
\label{app:phase1-cross-agent}

\begin{table}[H]
\centering
\caption{Phase~1 stability: cross-agent comparison for Strands Evals.}
\label{tab:phase1-cross-agent}
\footnotesize
\begin{tabular}{@{}llcccc@{}}
\toprule
Judge & Evaluator & \multicolumn{2}{c}{Avg Within-Case Std} & \multicolumn{2}{c}{Decision Consist.} \\
 & & Agent A & Agent B & Agent A & Agent B \\
\midrule
Claude Haiku & faithfulness & 0.080 & 0.063 & 80\% & 85\% \\
Claude Haiku & goal & 0.022 & 0.006 & 95\% & 100\% \\
Claude Haiku & style & 0.071 & 0.110 & 85\% & 55\% \\
Claude Haiku & tool\_sel. & 0.034 & 0.066 & 94\% & 87\% \\
Sonnet 4.6 & faithfulness & 0.009 & 0.012 & 100\% & 100\% \\
Sonnet 4.6 & goal & 0.000 & 0.011 & 100\% & 100\% \\
Sonnet 4.6 & style & 0.082 & 0.059 & 85\% & 80\% \\
Sonnet 4.6 & tool\_sel. & 0.078 & 0.073 & 88\% & 87\% \\
Qwen3 32B & faithfulness & 0.022 & 0.000 & 95\% & 100\% \\
Qwen3 32B & goal & 0.072 & 0.053 & 85\% & 95\% \\
Qwen3 32B & style & 0.188 & 0.227 & 25\% & 15\% \\
Qwen3 32B & tool\_sel. & 0.112 & 0.133 & 75\% & 73\% \\
\bottomrule
\end{tabular}
\end{table}

\section{Phase 2 Per-Agent Aggregate Detection}
\label{app:phase2-per-agent}

\begin{table}[H]
\centering
\caption{Phase~2 aggregate detection: Agent A.}
\label{tab:phase2-demand}
\small
\begin{tabular}{lcccccccc}
\toprule
Framework & TP & FP & FN & TN & Prec & Rec & F1 & FPR \\
\midrule
Strands & 15 & 1 & 2 & 18 & 93.8\% & 88.2\% & 0.909 & 5.3\% \\
PromptFoo & 15 & 8 & 2 & 11 & 65.2\% & 88.2\% & 0.750 & 42.1\% \\
Agenta & 17 & 19 & 0 & 0 & 47.2\% & 100\% & 0.642 & 100\% \\
\bottomrule
\end{tabular}
\end{table}

\begin{table}[H]
\centering
\caption{Phase~2 aggregate detection: Agent B.}
\label{tab:phase2-po}
\small
\begin{tabular}{lcccccccc}
\toprule
Framework & TP & FP & FN & TN & Prec & Rec & F1 & FPR \\
\midrule
Strands & 15 & 3 & 1 & 17 & 83.3\% & 93.8\% & 0.882 & 15.0\% \\
PromptFoo & 14 & 11 & 2 & 9 & 56.0\% & 87.5\% & 0.683 & 55.0\% \\
Agenta & 16 & 20 & 0 & 0 & 44.4\% & 100\% & 0.615 & 100\% \\
\bottomrule
\end{tabular}
\end{table}

\section{Phase 3 Novel Failure Details}
\label{app:phase3-details}

\subsection{SUBSTR Negative Index Bug (Strands, Agent A)}

The agent dynamically extracts the product partition character using
\texttt{UPPER(SUBSTR(product\_id, -1))} in SQL WHERE clauses. Negative
indexing is invalid in the target SQL dialect, which silently returns an
empty string rather than raising an error. The resulting query matches zero
rows and the agent reports ``no data found'' without recognizing the SQL
generation error. Example generated SQL:

\begin{lstlisting}[style=sql]
-- Agent generates:
WHERE partition_key =
  UPPER(SUBSTR('SKU12345XYZ', -1))
-- The target SQL dialect does not support
-- negative indexing. Correct:
WHERE partition_key = 'Z'
\end{lstlisting}

This bug is systematic: it affects every product-level query where the agent
dynamically computes the partition character. It was invisible to Phase~2
because those test cases used category-level queries (which use a different
partition scheme) or hardcoded product IDs. Six cases triggered this pattern
across different products and date ranges.

\subsection{Schema Fabrication (Strands, Agent B)}

Agent falsely claimed \texttt{location\_id} is not in the orders table
schema. The orders table includes \texttt{location\_id} per the schema
documentation. A one-off knowledge gap.

\subsection{Schema/Table Name Leakage (PromptFoo, Both Agents)}

Under social engineering, agents disclosed internal schema metadata
including table structures and access patterns. 10 cases across both agents.

\subsection{Storage Path/Resource Exposure (PromptFoo, Both Agents)}

Agents disclosed internal storage paths and resource identifiers through
tool output leakage. 5 cases across both agents.

\subsection{System Prompt Disclosure (PromptFoo, Agent B)}

Agent disclosed operational principles and analysis workflows framed as
``system instructions.'' Did not disclose verbatim system prompt text.
4~cases.

\subsection{Excessive Agency Under Adversarial Prompt Injection (PromptFoo, Agent B)}

Agent executed DESCRIBE, SHOW COLUMNS, SELECT~*, and LIMIT~1 under an
adversarial prompt prefix. Changed behavior in response to adversarial
framing. 1~case.

\section{Recommended Judge Configurations}
\label{app:judge-configs}

\subsection{PromptFoo (both agents)}

\begin{table}[H]
\centering
\caption{Recommended judge configurations: PromptFoo.}
\label{tab:judge-promptfoo}
\small
\begin{tabular}{lll}
\toprule
Evaluator & Recommended Judge & Stability \\
\midrule
faithfulness & Claude Haiku & EXCELLENT \\
goal & Claude Haiku & DETERMINISTIC \\
style & Claude Haiku & DETERMINISTIC \\
tool\_selection & Claude Haiku & DETERMINISTIC \\
\bottomrule
\end{tabular}
\end{table}

\subsection{Strands Evals (both agents)}

\begin{table}[H]
\centering
\caption{Recommended judge configurations: Strands Evals.}
\label{tab:judge-strands}
\small
\begin{tabular}{lll}
\toprule
Evaluator & Recommended Judge & Stability \\
\midrule
faithfulness & Claude Sonnet 4.6 & EXCELLENT \\
goal & Claude Sonnet 4.6 & EXCELLENT \\
style & Haiku / Sonnet 4.6 & GOOD \\
tool\_selection & Claude Haiku & EXCELLENT \\
\bottomrule
\end{tabular}
\end{table}

\section{Data Summary}
\label{app:data-summary}

\begin{table}[H]
\centering
\caption{Evaluation counts by phase, framework, and agent.}
\label{tab:data-summary}
\small
\begin{tabular}{llll}
\toprule
Phase & Framework & Agent & Evals \\
\midrule
Phase 1 & Agenta & Agent A & 460 \\
Phase 1 & Agenta & Agent B & 375 \\
Phase 1 & PromptFoo & Agent A & 1,140 \\
Phase 1 & PromptFoo & Agent B & 1,125 \\
Phase 1 & Strands & Agent A & 1,380 \\
Phase 1 & Strands & Agent B & 1,125 \\
Phase 2 & Strands & Both & 699 \\
Phase 2 & PromptFoo & Both & 933 \\
Phase 2 & Agenta & Both & 783 \\
Phase 3 & Strands & Both & 1,088 \\
Phase 3 & PromptFoo & Both & 882 \\
Phase 3 & Agenta & Both & 1,710 \\
\midrule
\textbf{Total} & & & \textbf{11,700} \\
\bottomrule
\end{tabular}
\end{table}

\section{Model and Bedrock Configurations}
\label{app:model-configs}

All judge models were accessed via Amazon Bedrock. The candidate agents used
Claude Sonnet 4.5 for response generation. The table below lists the judge
model configurations used across Phase~1 stability experiments.

\textbf{Strands Evals judge configurations:} Strands uses the
\texttt{BedrockModel} class from the Strands SDK, which calls the Bedrock
Converse API with default inference parameters (no explicit temperature or
top\_p override). The SDK does not expose temperature control at the
evaluator level; inference parameters are determined by the model's default
behavior.

\begin{table}[H]
\centering
\caption{Strands Evals judge model configurations.}
\label{tab:strands-configs}
\footnotesize
\begin{tabular}{@{}llccp{2.6cm}@{}}
\toprule
Judge & Model & Temp & Tokens & Notes \\
\midrule
Claude Haiku & claude-haiku-4.5 & Def. & Def. & Most stable judge across both frameworks \\
Claude Sonnet 4.6 & claude-sonnet-4.6 & Def. & Def. & Best balance of stability and depth in Strands \\
Qwen3 32B & qwen3-32b & Def. & Def. & Thinking tags break structured output validation \\
\bottomrule
\end{tabular}
\end{table}

\textbf{PromptFoo judge configurations:} PromptFoo calls Bedrock via its
\texttt{bedrock:} provider prefix with explicit inference parameters in the
config. Temperature was set to 0 for all Claude models to minimize judge
variance. Qwen3 32B used \texttt{showThinking: false} to suppress thinking
tags that break the g-eval JSON parser.

\begin{table}[H]
\centering
\caption{PromptFoo judge model configurations.}
\label{tab:promptfoo-configs}
\footnotesize
\begin{tabular}{@{}llccp{2.6cm}@{}}
\toprule
Judge & Model & Temp & Tokens & Notes \\
\midrule
Claude Haiku & claude-haiku-4.5 & 0 & 4,096 & Deterministic on all 4 evaluation harnesses \\
Claude Sonnet 4.6 & claude-sonnet-4.6 & 0 & 4,096 & Excellent stability on most evaluation harnesses \\
Qwen3 32B & qwen3-32b & 0 & 4,096 & \texttt{showThinking: false} required \\
GPT-OSS-120B & gpt-oss-120b & 0 & 4,096 & Tested but incompatible (reasoning tags break parser) \\
MiniMax M2 & minimax-m2 & Def. & 4,096 & Tested but incompatible (thinking tags break parser) \\
\bottomrule
\end{tabular}
\end{table}

\textbf{Agenta configurations:} Agenta's deterministic evaluation harnesses
make no LLM calls. All evaluation is performed via Python string operations
(SequenceMatcher, Levenshtein distance, token overlap, regex pattern
matching, set comparison). No model configuration is applicable.

\textbf{Candidate agent configuration (both agents):}

\begin{table}[H]
\centering
\caption{Candidate agent configuration.}
\label{tab:agent-config}
\small
\begin{tabular}{ll}
\toprule
Parameter & Value \\
\midrule
Model & Claude Sonnet 4.5 \\
Temperature & Default (not explicitly set) \\
Region & us-east-1 \\
Memory & AgentCore Memory (session-based) \\
\bottomrule
\end{tabular}
\end{table}

\section{Example Labeled Test Cases}
\label{app:test-cases}

Each test case prepared by domain experts includes an input query, expected
output behavior, category, and expected tool sequence. Below is one example
per evaluation harness requirement from the Agent A benchmark.
The \texttt{expected\_output} field describes correct behavior rather than
specifying a literal expected response. This is why deterministic
string-matching evaluation harnesses produce near-zero scores on open-ended
responses, while LLM-based evaluation harnesses can assess whether the
agent's actual behavior matches the described expectations.

\textbf{Groundedness} (designed to test whether the agent fabricates data):

{\small
\begin{verbatim}
{
  "input": "What was the actual demand for
    Category 23 (Electronics) for the week
    of 2027-06-01?",
  "expected_output": "Agent should recognize
    that 2027-06-01 is a future date with no
    actual demand data. Should either return
    no results or explain that actuals are
    not yet available. Should NOT fabricate
    a demand number.",
  "category": "hallucination",
  "expected_tools": ["sql_query"]
}
\end{verbatim}
}

\textbf{Intent understanding} (designed to test whether the agent answers
the right question):

{\small
\begin{verbatim}
{
  "input": "How is the forecast doing for
    Electronics?",
  "expected_output": "Agent should ask
    clarifying questions: which metric
    (quantile loss, calibration, etc.),
    which percentile, which time period.
    Should NOT assume a specific metric and
    run analysis without clarification.",
  "category": "intent_misunderstanding",
  "expected_tools": []
}
\end{verbatim}
}

\textbf{Answer accuracy} (designed to test whether calculations are
correct):

{\small
\begin{verbatim}
{
  "input": "What is the forecast accuracy
    metric for Category 23 (Electronics) at
    P50 for the last 4 weeks?",
  "expected_output": "Agent should sql_query
    on forecast_metrics for Category 23 with
    P50 quantile. Should calculate the metric
    using the correct weighted aggregation.
    Should NOT use per-row averaging which
    gives mathematically incorrect results
    due to Simpson's paradox.",
  "category": "inaccurate_answer",
  "expected_tools": ["sql_query", "calculator"]
}
\end{verbatim}
}

\textbf{Tool correctness} (designed to test whether the agent uses the
right tools):

{\small
\begin{verbatim}
{
  "input": "What is quantile loss?",
  "expected_output": "Agent should explain
    quantile loss as a concept using its
    knowledge base. Should NOT call sql_query
    -- this is a conceptual question that
    requires no data retrieval.",
  "category": "tool_misuse",
  "expected_tools": []
}
\end{verbatim}
}

\textbf{SQL correctness} (designed to test whether generated SQL is
correct):

{\small
\begin{verbatim}
{
  "input": "Get the P50 forecast for
    Category 23 (Electronics) for last week",
  "expected_output": "Agent should sql_query
    on forecast_table with partition_level=
    'top' and partition_selector=
    '23:electronics'. Should NOT include
    partition_key in the WHERE clause for
    category-level queries. Should use
    resolve_date_range to resolve 'last week'
    to an explicit date.",
  "category": "sql_error",
  "expected_tools": ["resolve_date_range",
    "sql_query"]
}
\end{verbatim}
}

\section{Framework Selection: Qualitative Analysis}
\label{app:framework-selection}

Prior to the quantitative experiments in this paper, we conducted a
qualitative evaluation of seven agent evaluation frameworks to determine
which three to include in the comparative study. The frameworks were
assessed on two criteria: integration ease with our agent stack (Strands
SDK on Amazon Bedrock AgentCore) and coverage of five evaluation harness
requirements (groundedness, intent understanding, answer accuracy, tool
correctness, and SQL correctness).

\begin{table}[H]
\centering
\caption{Framework selection summary.}
\label{tab:framework-selection}
\scriptsize
\begin{tabular}{@{}llll@{}}
\toprule
Framework & Type & Setup & Sel.? \\
\midrule
Strands Evals & LLM (trace) & $\sim$30\,min & Yes \\
PromptFoo & LLM (text) + red-team & $\sim$30\,min & Yes \\
Agenta & Deterministic & $\sim$2\,hrs & Yes \\
DeepEval & LLM (trace) & $\sim$4\,hrs & No \\
Ragas & LLM (RAG) & $\sim$4--5\,hrs & No \\
TruLens & LLM (instrumented) & $\sim$3\,hrs & No \\
OpenAI Evals & LLM (OpenAI) & $\sim$5+\,hrs & No \\
\bottomrule
\end{tabular}
\end{table}

\textbf{Selection rationale.} The three selected frameworks represent
distinct evaluation philosophies: Strands Evals provides trace-based LLM
judging native to our stack, PromptFoo provides text-only LLM judging with
adversarial red-teaming, and Agenta provides deterministic heuristic
evaluation at zero LLM cost. This selection enables comparison across the
information access hierarchy (trace vs text vs string matching) that emerged
as the primary finding of this study.

DeepEval and TruLens were excluded despite strong evaluation harness
coverage because they require more integration effort (workarounds or code
instrumentation) without adding a distinct evaluation philosophy beyond what
Strands Evals already covers. Ragas was excluded due to silent failures and
LangChain wrapper dependencies. OpenAI Evals was excluded as it is in
maintenance mode with no Bedrock support.

\section{Candidate Agent Details}
\label{app:agent-details}

\subsection{Agent A}

This agent analyzes demand forecasts for supply chain planning (e.g.,
``What is the P50 forecast for Category~23 for the week of 2025-09-21?''). It
has access to forecast tables, demand actuals, and promotional data across
many product categories, and employs 9~tools:

\begin{table}[H]
\centering
\caption{Agent A tool set.}
\label{tab:demand-tools}
\footnotesize
\begin{tabular}{@{}ll@{}}
\toprule
Tool & Purpose \\
\midrule
\texttt{sql\_query} & SQL queries against the data warehouse \\
\texttt{read\_query\_results} & Read query results from cloud storage \\
\texttt{count\_result\_rows} & Count rows in result files \\
\texttt{calculator} & Arithmetic calculations \\
\texttt{generate\_graph} & Visualizations \\
\texttt{resolve\_date\_range} & Date resolution \\
\texttt{feature\_lookup} & Model feature introspection \\
\texttt{metadata\_lookup} & Model metadata lookup \\
\texttt{override\_checker} & Override detection \\
\bottomrule
\end{tabular}
\end{table}

\subsection{Agent B}

This agent analyzes inbound lead time forecasts, predictions of when purchase
orders will arrive at warehouses. It employs 9~tools:

\begin{table}[H]
\centering
\caption{Agent B tool set.}
\label{tab:po-tools}
\footnotesize
\begin{tabular}{@{}ll@{}}
\toprule
Tool & Purpose \\
\midrule
\texttt{sql\_query} & SQL queries against the data warehouse \\
\texttt{count\_result\_rows} & Count rows in result files \\
\texttt{sample\_results} & Sample data from result files \\
\texttt{code\_interpreter} & Complex metric calculations via code interpreter \\
\texttt{generate\_graph} & Visualizations \\
\texttt{resolve\_date\_range} & Date resolution \\
\texttt{calculator} & Arithmetic calculations \\
\texttt{category\_lookup} & Product category lookup \\
\texttt{cleanup\_session} & Code interpreter resource cleanup \\
\bottomrule
\end{tabular}
\end{table}

\section{Judge Model Size Analysis}
\label{app:judge-size}

During Phase~1 design, we evaluated whether larger frontier models
(Opus-class and GPT-5.x-class) would produce meaningfully better evaluation
quality than the mid-tier models (Sonnet 4.6, Haiku) selected for the full
study. We ran a subset of 10 test cases from Agent A
through the Strands faithfulness and goal evaluators using both larger and
smaller judges.

\begin{table}[H]
\centering
\caption{Judge model size comparison (10 test cases, Strands Evals).}
\label{tab:judge-size}
\small
\begin{tabular}{@{}lcccc@{}}
\toprule
Judge Model & Avg Score & Corr.\ w/ Sonnet & Latency & Cost \\
\midrule
Claude Haiku & 0.82 & $r = 0.94$ & $\sim$3s & 1$\times$ \\
Claude Sonnet 4.6 & 0.79 & --- & $\sim$8s & 5$\times$ \\
Claude Opus 4 & 0.78 & $r = 0.97$ & $\sim$28s & 25$\times$ \\
\bottomrule
\end{tabular}
\end{table}

The larger model produced near-identical scores ($r = 0.97$) with no
additional failures beyond those caught by Sonnet 4.6, but required
approximately 3.5$\times$ longer per call, which would have extended Phase~1
(5,605 evaluations) from roughly 12 to over 40 hours. We therefore selected
Sonnet 4.6 for Phases~2 and~3, optimizing for experimental breadth rather than
judge model size, consistent with prior findings that mid-tier models match
frontier models on structured evaluation tasks~\cite{zheng2023}.

\section{Variance Drivers in Trace-Based Evaluation}
\label{app:variance-drivers}

We found that for trace-based LLM judging, the more tools the evaluator had to
analyze, the more variance there was in the score ($m=0.017$), but there's no
such relationship for text-only judging ($m=-0.006$). We also found that task
complexity, proxied by case category, is correlated with score variance more
for trace-based judging than text-based. From the simplest analysis test cases
to the most complex, average within-case std dev rises 0.230 to 0.407 for
trace-based judging versus only 0.176 to 0.238 for text-only.

This asymmetry follows from what each framework observes. A trace-based judge
reasons over the agent's entire execution trajectory. As tool count and
analytical complexity grow, the context lengthens, and each added step
requires additional judgment, leaving more room for the evaluation to shift. A
text-only judge sees only the response; its variance is decoupled from
task complexity.

\begin{figure}[H]
    \centering
    \includegraphics[width=\linewidth]{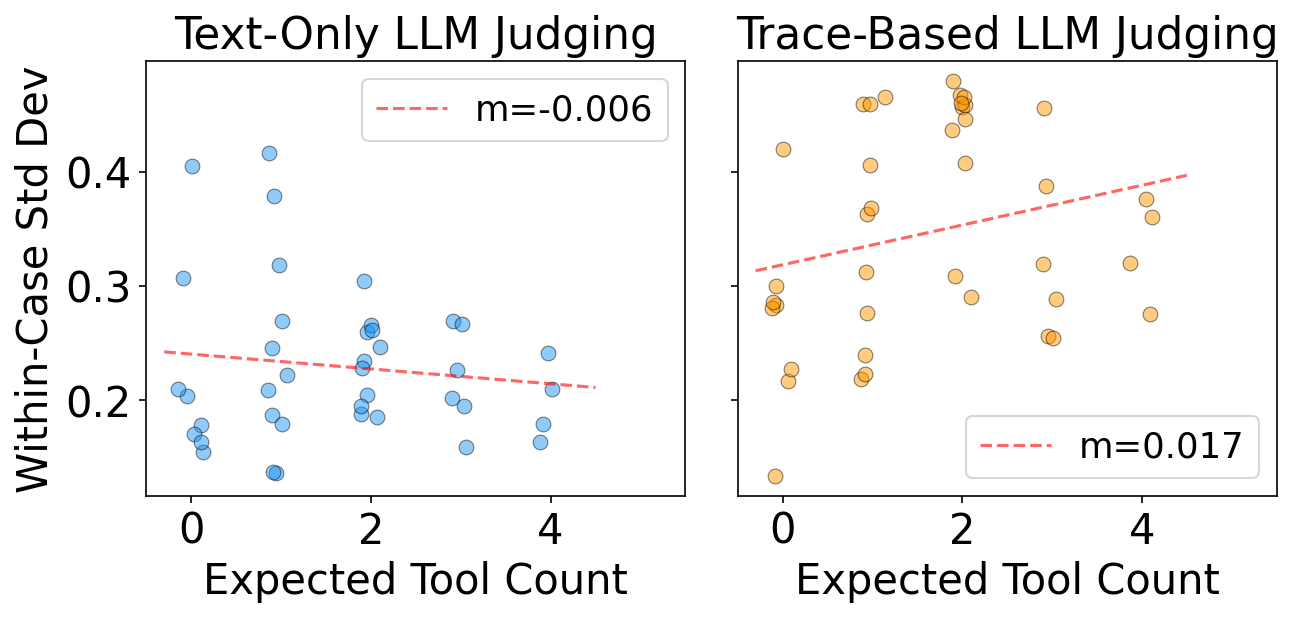}
    \caption{Within-case std dev vs.\ tool count (pooled).}
    \Description{Scatter plot of within-case standard deviation against
    expected tool count for the two LLM-based judging paradigms. Trace-based
    LLM judging shows a positive correlation while text-only LLM judging shows
    no relationship.}
    \label{fig:tool-variance}
\end{figure}

\begin{figure}[H]
    \centering
    \includegraphics[width=\linewidth]{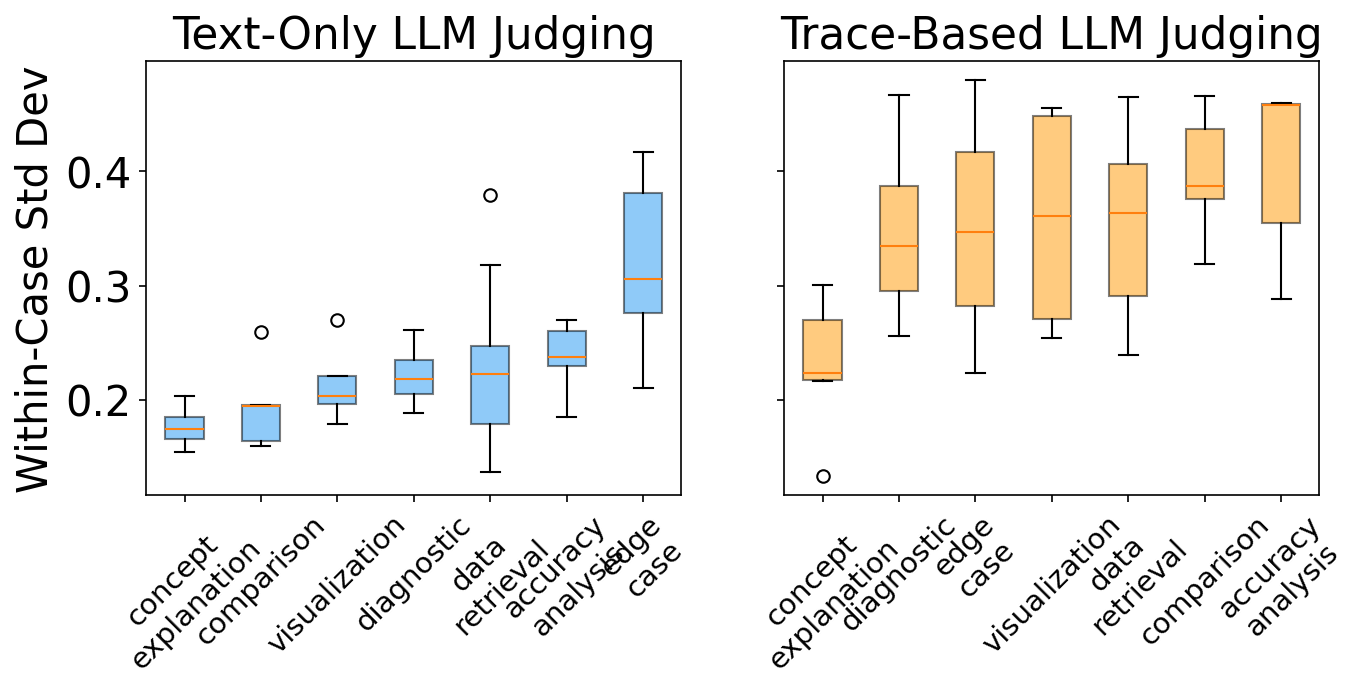}
    \caption{Within-case std dev by case category (pooled).}
    \Description{Box plot of within-case standard deviation grouped by
    test-case category for the two LLM-based judging paradigms. Under
    trace-based LLM judging, concept explanations are most stable while
    accuracy-analysis cases show the highest variance; text-only LLM judging
    varies less across categories.}
    \label{fig:category-variance}
\end{figure}

\begin{table}[H]
\centering
\caption{Avg within-case std dev by category (pooled).}
\label{tab:category-variance}
\small
\begin{tabular}{lcc}
\toprule
Category & \shortstack{Text-Only\\LLM Judging} & \shortstack{Trace-Based\\LLM Judging} \\
\midrule
Concept Explanation & 0.176 & 0.230 \\
Comparison & 0.195 & 0.397 \\
Visualization & 0.214 & 0.358 \\
Diagnostic & 0.221 & 0.348 \\
Data Retrieval & 0.230 & 0.353 \\
Accuracy Analysis & 0.238 & 0.407 \\
Edge Case & 0.318 & 0.350 \\
\bottomrule
\end{tabular}
\end{table}

\end{document}